\documentclass[conference]{IEEEtran}
\IEEEoverridecommandlockouts

\usepackage{cite}
\usepackage{amsmath,amssymb,amsfonts}
\usepackage{algorithmic}
\usepackage{graphicx}
\usepackage{textcomp}
\usepackage{xcolor}
\usepackage{balance}
\def\BibTeX{{\rm B\kern-.05em{\sc i\kern-.025em b}\kern-.08em
    T\kern-.1667em\lower.7ex\hbox{E}\kern-.125emX}}
\begin{document}
\bstctlcite{IEEEexample:BSTcontrol}

\title{Open-Source Factor Graph Optimization Package for GNSS: Examples and Applications\\
}

\author{\IEEEauthorblockN{1\textsuperscript{st} Taro Suzuki}
\IEEEauthorblockA{\textit{Future Robotics Technology Center,} \\
\textit{Chiba Institute of Technology}\\
Chiba, Japan \\
taro@furo.org}
}

\maketitle

\begin{abstract}
State estimation methods using factor graph optimization (FGO) have garnered significant attention in global navigation satellite system (GNSS) research. FGO exhibits superior estimation accuracy compared with traditional state estimation methods that rely on least-squares or Kalman filters. However, only a few FGO libraries are specialized for GNSS observations. This paper introduces an open-source GNSS FGO package named gtsam\_gnss, which has a simple structure and can be easily applied to GNSS research and development. This package separates the preprocessing of GNSS observations from factor optimization. Moreover, it describes the error function of the GNSS factor in a straightforward manner, allowing for general-purpose inputs. This design facilitates the transition from ordinary least-squares-based positioning to FGO and supports user-specific GNSS research. In addition, gtsam\_gnss includes analytical examples involving various factors using GNSS data in real urban environments. This paper presents three application examples: the use of a robust error model, estimation of integer ambiguity in the carrier phase, and combination of GNSS and inertial measurements from smartphones. The proposed framework demonstrates excellent state estimation performance across all use cases.
\end{abstract}


\section{Introduction}
Improving the positioning accuracy of global navigation satellite systems (GNSS) in urban environments remains a key challenge in navigation research \cite{urbannav}. Research on high-precision positioning in urban environments can be divided into two categories: methods for reducing positioning errors owing to multipath effects caused by signal reflection and diffraction, and methods for integrating GNSS data with other sensors and data to improve performance in environments where GNSS is temporarily unavailable or the number of available satellites is limited. Multipath mitigation methods typically focus on the identification and elimination of non-line-of-sight (NLOS) multipath effects, where only reflected and diffracted signals are received from obstructed satellites \cite{nlos_general}. In addition, researchers have explored the integration of GNSS with inertial measurement units (IMUs), cameras, lidar, and 3D maps. Active research is being conducted to improve the performance and reliability of GNSS in urban environments \cite{gpsins}.

Traditionally, GNSS positioning has been based on the least-squares method, and Bayesian filters such as the Kalman filter (KF) have been applied to reduce errors and integrate other sensors and data sources \cite{gnss_handbook}. Position estimation using KF is embedded into GNSS receivers and contributes to improving positioning accuracy in urban environments. Moreover, in recent years, optimization-based state estimation methods, such as factor graph optimization (FGO), have been widely used in GNSS research \cite{gnssfgo_2012, gnssfgo_survey}.

The increased interest in optimization-based state estimation is attributable to its superior performance (higher state-estimation accuracy) compared with conventional filtering methods such as extended KF \cite{gnssfgo_kf}. Although optimization-based state estimation methods are computationally expensive, their practicality has improved with recent advancements in computing power.

FGO was originally developed for robotics and has been widely used for robot localization and mapping using sensors such as lidar, cameras, and IMUs \cite{fgo_general1,fgo_book}. Various open-source FGO-based robot position-estimation packages are currently available \cite{orbslam,cartographer}. Recently, open-source FGO packages that use GNSS observations for state estimation have been established for GNSS research \cite{robustgnss,graphgnsslib}. However, these FGO packages are large-scale and exhibit a complex program structure, rendering them difficult to adapt for specific GNSS problem settings and research. In addition, the estimated states are often implemented in original classes, which makes it challenging to appropriately respond to changes in the type of GNSS observations used or estimated states.

To address these limitations, this paper introduces a simple FGO package, gtsam\_gnss, which can be easily applied to research and development using actual GNSS data. This package is available as an open-source project\footnote{\texttt{https://github.com/taroz/gtsam\_gnss}}. This package facilitates the implementation of FGO using GNSS observations. It can be easily adapted to specific GNSS use cases by creating custom factors or modifying the combination of factors. In addition, this package can be used for educational purposes, such as training or improving optimization-based strategies for GNSS positioning.

\subsection{Related Studies}
FGO is widely used for robot state estimation, and various open-source libraries for FGO processing, such as g$^2$o \cite{g2o}, Ceres Solver \cite{ceres}, and GTSAM \cite{gtsam}, have been released. These libraries use Lie group-based 3D pose expressions to represent the state and perform optimization on Lie algebras. Numerous open-source FGO-based position estimation methods that use these libraries as backends have been released \cite{fgo_general1,fgo_book,orbslam,cartographer}.

Notably, most of the existing FGO-based position estimation methods use lidar or camera observations as factors. When GNSS is used as a factor, many methods directly use the position output from the GNSS receiver as a factor in the position estimation \cite{cartographer,hdlslam,liosam}. In GTSAM \cite{gtsam}, the 3D position constraint is provided as a GPS factor, and in studies \cite{cartographer}, \cite{hdlslam}, and \cite{liosam}, the GNSS position constraint is incorporated into the lidar odometry using FGO. This method, termed loose coupling, uses the position computed from the GNSS observations as the constraint. If the GNSS receiver cannot output a position owing to a decrease in the number of satellites, GNSS observations cannot be used for optimization. In addition, if the estimated position is degraded owing to multipath effects, the optimization accuracy is considerably deteriorated.

In contrast, several open-source FGO packages use raw GNSS observations, a strategy termed tight coupling \cite{robustgnss,graphgnsslib,gnssfgo,posgo,gicilib}. For example, the approach presented in \cite{robustgnss} uses GTSAM as a backend and incorporates only pseudorange observations. Other studies \cite{gvins,posgo} provides factors for GNSS Doppler observations in addition to pseudorange observations. Other frameworks, such as \cite{graphgnsslib}, \cite{gnssfgo}, and \cite{gicilib}, use factors that incorporate not only pseudorange and Doppler observations, but also carrier-phase measurements for high-precision positioning. The open-source packages in \cite{gnssfgo} and \cite{gicilib} combine GNSS with IMUs, lidar, and cameras. Notably, \cite{gicilib} estimates the integer ambiguity of the carrier phase and achieves high-precision positioning. However, the architecture of this program is complex, and it relies on the robot operating system (ROS), a middleware for robots, requiring substantial effort for optimizing GNSS alone or incorporating new factors or custom graph structures based on user-specific GNSS data. Although \cite{graphgnsslib} provides a simple open-source graph optimization library specialized for GNSS, it relies on ROS and is thus difficult to apply for GNSS positioning in a non-ROS environment.

Overall, the existing open-source GNSS FGO packages are specialized for specific applications and have a complex structure, rendering it challenging for beginners to learn and use them for custom problem settings with modified inputs and graph structures. Therefore, this paper introduces a novel open-source GNSS FGO package that is simple and user-friendly.

\subsection{Contributions}
The features and contributions of the proposed package can be summarized as follows.

\begin{itemize}
\item By separating the processing of GNSS observations from the optimization process, the optimization process can be implemented in a straightforward manner. This design allows the optimization process to have general-purpose inputs, facilitating the transition from conventional least-squares-based positioning to FGO.

\item The proposed package supports individual factors based on combinations of multiple GNSS observations (pseudorange, Doppler, and carrier phase) and different estimation states (e.g., position, velocity, receiver clock, receiver clock drift, and carrier-phase ambiguity), allowing users to select the appropriate combination of factors for specific problems.

\item Processing of GNSS observations can be performed using MATLAB, and the package provides wrappers for factors of GNSS observations that can be called from MATLAB. This makes it possible to establish an FGO pipeline using MATLAB, allowing beginners to easily learn the package and use it for developing and verifying prototypes in GNSS research.
\end{itemize}

\section{FGO Model for GNSS Observations}
\subsection{Overview}
The proposed open-source FGO package for GNSS, gtsam\_gnss, uses the general-purpose FGO library GTSAM \cite{gtsam} as its backend. GTSAM is an open-source library written in C++ that provides state and factor classes, which are crucial for graph optimization, as well as several types of optimization solvers. Additionally, GTSAM provides various robust error models for M-estimators, enabling robust optimization. In GNSS positioning, where multipath errors and cycle slips in pseudorange and carrier-phase observations as outliers can significantly degrade accuracy, robust optimization is highly beneficial.

\begin{figure}[t]
    \centerline{\includegraphics[width=88mm]{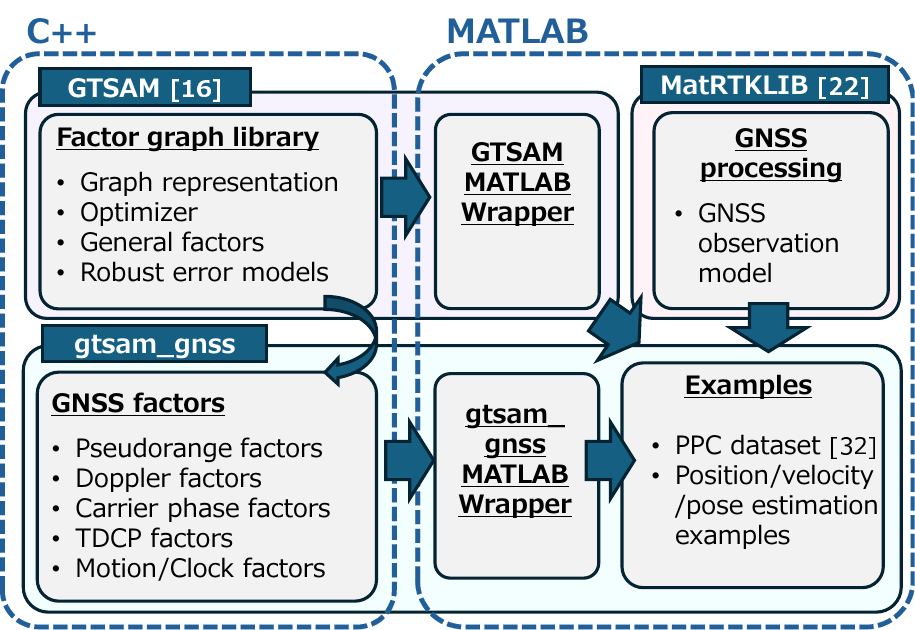}}
    \caption{Software structure of gtsam\_gnss. gtsam\_gnss provides GNSS observation factors written in C++ and their MATLAB wrappers. Additionally, it includes datasets and analytical examples and can be implemented with the GTSAM optimization library and MatRTKLIB GNSS processing library.}
    \label{fig1}
\end{figure}

Fig. 1 shows an overview of the gtsam\_gnss package. The package implements factors for different GNSS observations as independent source code written in C++ using the GTSAM factor templates. In addition, it provides wrappers to call these factors from MATLAB.
Notably, gtsam\_gnss separates the preprocessing of GNSS observations from the FGO process. For example, several process are typically required to compute the pseudorange residuals for position estimation, such as satellite position computation, geometric distance computation, and ionospheric and tropospheric delay correction. gtsam\_gnss adopts an open-source MATLAB-based GNSS analysis library named MatRTKLIB, developed by the author of this paper \cite{matrtklib}. By separating the GNSS preprocessing and optimization processes, the error function within the factor can be easily defined. This allows the provided factors to have generic inputs, and thus, the factors provided by gtsam\_gnss can be conveniently adapted to different configuration problems.

The gtsam\_gnss package provides various factors based on GNSS observations, along with specific examples of analyses using these factors. For example, it includes factors for estimating (1) position and receiver clock using basic pseudorange observations, (2) velocity and receiver clock drift using Doppler, (3) both position and velocity, (4) relative position using time-differenced carrier phase (TDCP), and (5) ambiguity using double-differenced (DD) carrier-phase observations. By separating the estimated state into individual variables such as position, velocity, and clock, users can select the appropriate combination of factors and estimated states in accordance with the problem settings. This section outlines the factor models for GNSS observations used in gtsam\_gnss.

\subsection{Pseudorange Factor}
The GNSS pseudorange factor can be used for position estimation. GNSS pseudorange observations include receiver clock errors, ionospheric and tropospheric errors, and satellite orbit and clock errors. Typically, for single-point positioning, receiver clock errors (and inter-satellite system bias) are incorporated in the estimated state in addition to the position. However, when using the difference in pseudoranges between satellites, the receiver clock error is canceled. Therefore, gtsam\_gnss provides two types of pseudorange factors, one with receiver clock included in the estimated state and one without.

To enhance the linearization and optimization efficiencies, gtsam\_gnss uses the error state of the 3D position $\delta \mathbf{x}$ with respect to the initial position $\mathbf{x}_0$ as the estimated state. The relation between the $i$-th state and error state is $\mathbf{x}_i=\mathbf{x}_{0}+\delta\mathbf{x}_i$. Let $\rho_i^{k}$ denote the GNSS pseudorange from satellite $k$ at the $i$-th epoch. The error function of the pseudorange factor is expressed as follows:

\begin{equation}
    e_{\mathrm{P}}^k\left(\delta\mathbf{x}_i,\mathbf{c}_i\right)=\mathbf{H}_{\mathbf{x}_0}^k \delta \mathbf{x}_i + \mathbf{H}_{\mathbf{c}}^k \mathbf{c}_i -\left(\rho_i^k-r^{k}_{\mathbf{x}_0,i}-\varepsilon_i^k\right)
\end{equation}

\noindent where $\mathbf{c}_i$ denotes the receiver clock error and inter-satellite system bias, $r^{k}_{\mathbf{x}_0,i}$ is the geometric distance between satellite $k$ and the initial position ${\mathbf{x}_0}$, and $\varepsilon_{i}^{k}$ denotes the pseudorange correction value including ionospheric and tropospheric errors. $\varepsilon_{i}^{k}$ is estimated using a general GNSS observation model (e.g., the Klobuchar model for ionospheric errors) for single-point positioning, whereas pseudorange observations of a reference station are used for differential positioning. $\mathbf{H}_{\mathbf{x}_0}^{k}$ is the line-of-sight (LOS) vector to the satellite at the initial position $\mathbf{x}_{0}$, and $\mathbf{H}_{\mathbf{c}}^k$ is the observation matrix determined by the satellite system used. These parameters can be expressed as follows:

\begin{equation}
    \mathbf{H}_{\mathbf{x}_0}^{k}=\left[\begin{array}{llll}
    {u_{x,i}^k} & {u_{y,i}^k} & {u_{z,i}^k}
    \end{array}\right]
\end{equation}

\begin{equation}
    \mathbf{H}_{\mathbf{c}}^{k}=\left[\begin{array}{llllll}
    {1} & {\delta_{\mathrm{glo}}^{k}} & {\delta_{\mathrm{gal}}^{k}} & {\delta_{\mathrm{bds}}^{k}} & \cdots
    \end{array}\right]
\end{equation}

\noindent where $u_{x,i}^k$, $u_{y,i}^k$, and $u_{z,i}^k$ denote the components of the unit LOS vector to satellite $k$. $\delta_{\mathrm{glo}}^{k}$, $\delta_{\mathrm{gal}}^{k}$, and $\delta_{\mathrm{bds}}^{k}$ are equal to “1” when the $k$-th GNSS measurement is from GLONASS, Galileo, or BeiDou, and “0” otherwise. In addition, if observations from other satellite systems or other frequencies are used, the estimated states are added to $\mathbf{c}_i$ and $\mathbf{H}_{\mathbf{c}}^k$.

The last term in (1) represents the pseudorange residual at the initial position $\mathbf{x}_0$, which can be calculated a priori. By calculating the pseudorange residuals at the initial position and satellite LOS vector $\mathbf{H}_{\mathbf{x}_0}^k$ in advance and using them as inputs to the factor, the optimization process is simplified and becomes more versatile. The coordinate system of the estimated position can be adjusted as needed. Specifically, the earth-centered earth-fixed (ECEF) coordinate system can be used if $\mathbf{H}_{\mathbf{x}_0}^k$ is input in ECEF coordinates, or the east-north-up (ENU) coordinate system can be used if $\mathbf{H}_{\mathbf{x}_0}^k$ is input in the ENU coordinates. This flexibility is especially beneficial when compositing with IMUs, which typically operate in the ENU coordinate system.

The pseudorange factor when the receiver clock is not estimated, such as when calculating the difference in pseudoranges between satellites, can be defined as follows:

\begin{equation}
    e_{\mathrm{P}}^{kl}\left(\delta\mathbf{x}_i\right)=\mathbf{H}_{\mathbf{x}_0}^{kl} \delta \mathbf{x}_i  -\left(\rho_i^{kl}-r^{kl}_{\mathbf{x}_0,i}-\varepsilon_i^k\right)
\end{equation}

\noindent where $l$ represents the reference satellite.

gtsam\_gnss provides both types of pseudorange factors, enabling position estimation using GNSS pseudorange observations in problems such as single-point and differential positioning.

\subsection{Doppler Factor}
The 3D velocity $\mathbf{v}$ can be estimated from GNSS Doppler observations. Although the Doppler observations include a receiver clock drift ${d}$, this drift can be eliminated using the Doppler difference between satellites. Similar to position estimation, the estimated state is defined as the error state $\delta \mathbf{v}$ with respect to the initial velocity $\mathbf{v}_0$. Let $\Dot \rho_i^{k}$ denote the pseudorange rate converted from the Doppler observation from satellite $k$. The following two Doppler factors are provided to estimate the 3D velocity.

\begin{equation}
    e_{\mathrm{D}}^k\left(\delta\mathbf{v}_i,{d}_i\right)=\mathbf{H}_{\mathbf{x}_0}^k \delta \mathbf{v}_i + {d}_i - \left( \Dot \rho_i^{k} - \mathbf{H}_{\mathbf{x}_0}^k \mathbf{v}_{\mathrm{s},i}^{k}-\Dot{t}_{\mathrm{s},i}^k \right)
\end{equation}

\begin{equation}
    e_{\mathrm{D}}^{kl}\left(\delta\mathbf{v}_i\right)=\mathbf{H}_{\mathbf{x}_0}^{kl} \delta \mathbf{v}_i  - \left( \Dot \rho_i^{kl} - \mathbf{H}_{\mathbf{x}_0}^{kl} \mathbf{v}_{\mathrm{s},i}^{kl}-\Dot{t}_{\mathrm{s},i}^{kl} \right)
\end{equation}

\noindent where $\mathbf{v}_{\mathrm{s},i}^k$ is the satellite velocity, and $\Dot{t}_{\mathrm{s},i}^k$ is the satellite clock drift computed from the satellite ephemeris. Equation (5) defines the normal Doppler factor, and (6) is used when inter-satellite Doppler difference observations are used.

gtsam\_gnss provides two additional factors when Doppler observations are used as a constraint for successive 3D position estimates. If $\Delta$ as the operator that calculates the time difference, the time difference in the error states of the 3D position can be expressed as $\Delta \delta\mathbf{x}_{i} = \delta\mathbf{x}_{i}-\delta\mathbf{x}_{i-1}$. Let $\bar{\Dot \rho}_i^{k}$ denote the average Doppler observation between epochs. The error functions are expressed as follows:

\begin{IEEEeqnarray}{lCr}
    e_{\mathrm{D}}^k\left(\Delta\delta\mathbf{x}_{i},\Delta\mathbf{c}_{i}\right)=&\mathbf{H}_{\mathbf{x}_0}^k \Delta \delta \mathbf{x}_{i} + \mathbf{H}_{\mathbf{c}}^k \Delta \mathbf{c}_{i} \nonumber\\ &- \Delta t \left( \bar{\Dot \rho}_i^{k} - \mathbf{H}_{\mathbf{x}_0}^{k} \bar{\mathbf{v}}_{\mathrm{s},i}^{k}-\bar{\Dot{t}}_{\mathrm{s},i}^{k} \right)
\end{IEEEeqnarray}

\begin{IEEEeqnarray}{lCr}
    e_{\mathrm{D}}^k\left(\Delta\delta\mathbf{x}_{i}\right)=\mathbf{H}_{\mathbf{x}_0}^k \Delta \delta \mathbf{x}_{i} - \Delta t \left( \bar{\Dot \rho}_i^{k} - \mathbf{H}_{\mathbf{x}_0}^{k} \bar{\mathbf{v}}_{\mathrm{s},i}^{k}-\bar{\Dot{t}}_{\mathrm{s},i}^{k} \right)
\end{IEEEeqnarray}

\noindent where $\Delta t$ is the time step, (7) defines the normal Doppler observation, and (8) is used when the Doppler difference between satellites is considered. Instead of estimating velocity directly, these factors can be used to improve the performance of position estimation using pseudorange factors, or they can be used to estimate the relative position.

As shown in (5) to (8), gtsam\_gnss provides four types of Doppler factors, and these factors can be selected according to the estimated state and specific problem.

\subsection{TDCP Factor}
The TDCP factor can be used to estimate relative positions using GNSS carrier-phase observations. Although Doppler observations can also be used to estimate the relative position, TDCP provides significant higher accuracy \cite{vel1,td1}. Let TDCP observation $\Delta {\Phi}_{i}^{k}$ between sequential epochs $i-1$ and $i$ can be expressed as follows: 

\begin{equation}
    \lambda \Delta \Phi_{i}^{k} = \lambda \left[\Phi_{i}^{k}-\Phi_{i-1}^{k} \right] 
    \simeq \Delta r_{\mathbf{x},i}^{k} + \Delta t_{\mathrm{s},i}^{k} + \Delta c_{i}
\end{equation}

\noindent where $\lambda$, $\Phi_{i}^{k}$, and $ t_{\mathrm{s},i}^{k}$ denote the signal wavelength, measured carrier phase in cycles, and satellite clock error, respectively. When the time difference $\Delta t$ is small, the ionospheric and tropospheric delays and satellite orbit errors in the carrier phase are effectively canceled. 

As in the previous discussion, by differencing TDCP observations between satellites, the receiver clock change $\Delta c_{i}$ in the TDCP observations can be canceled. gtsam\_gnss provides the following two types of TDCP factors:

\begin{IEEEeqnarray}{lCr}
    e_{\mathrm{T}}^k\left(\Delta\delta\mathbf{x}_{i},\Delta\mathbf{c}_{i}\right)=&\mathbf{H}_{\mathbf{x}_0}^k \Delta\delta \mathbf{x}_{i} \nonumber+ \mathbf{H}_{\mathbf{c}}^k \Delta\mathbf{c}_{i}\\& - \left( \lambda \Delta {\Phi}_{i}^{k} - \Delta r^{k}_{\mathbf{x}_{0},i} - \Delta t_{\mathrm{s},i}^{k} \right)
\end{IEEEeqnarray}

\begin{IEEEeqnarray}{lCr}
    e_{\mathrm{T}}^{kl}\left(\Delta\delta\mathbf{x}_{i}\right)=&\mathbf{H}_{\mathbf{x}_0}^{kl} \Delta\delta \mathbf{x}_{i} & - \left( \lambda \Delta {\Phi}_{i}^{kl} - \Delta r^{kl}_{\mathbf{x}_{0},i} - \Delta t_{\mathrm{s},i}^{kl} \right)
\end{IEEEeqnarray}

Although TDCP is highly accurate, its availability is lower than that of Doppler observations owing to cycle slips and half-cycle ambiguities. If the carrier phase with cycle slip is not eliminated, the TDCP factor introduces relative position estimation errors. Thus, when using the TDCP factor, a robust error model should be used to mitigate the adverse effects of cycle slips \cite{ral2022_taro,fgo_tdcp}.

\subsection{Carrier Phase Factor}
Double-differenced (DD) carrier-phase observations, which refer to inter-reference station and inter-satellite observations, are typically used for high-precision positioning, such as real-time kinematic GNSS. In DD carrier-phase observations, most of the errors caused by the satellite, receiver, and signal propagation are eliminated. However, the integer ambiguity associated with the carrier phase remains. Defining $\nabla \Delta$ as an operator that calculates the double differentiation, the observation model for the DD carrier phase can be expressed as follows:

\begin{equation}
    \lambda \nabla \Delta \Phi_{i}^{kl} =\nabla \Delta r_{\mathbf{x},i}^{kl}+\lambda \nabla \Delta B_{i}^{kl}+\nabla \Delta \epsilon_{i}^{kl}
\end{equation}

\noindent where $\nabla \Delta B_{i}^{kl}$ represents the DD integer ambiguity, and $\epsilon_{i}^{kl}$ denotes the carrier-phase measurement noise. 

In the carrier-phase factor, the 3D position and carrier-phase ambiguity are estimated using the DD carrier-phase observation residuals. The carrier-phase factor can be defined as

\begin{IEEEeqnarray}{lCr}
    e_{\mathrm{B}}^k\left(\delta\mathbf{x}_{i}, \nabla \Delta B_{i}^{kl}\right)=\mathbf{H}_{\mathbf{x}_0}^{kl} \delta \mathbf{x}_{i} \nonumber\\ {}-\left(\lambda \nabla \Delta \Phi_{i}^{kl}-\nabla \Delta r^{kl}_{\mathbf{x}_0,i}-\lambda \nabla \Delta B_{i}^{kl}\right)
\end{IEEEeqnarray}

\noindent where the carrier-phase ambiguity is estimated as a float number during optimization. Notably, techniques such as the integer least-squares method \cite{ar_review} must be applied to the float ambiguity estimated using the FGO to resolve the integer ambiguity.

\subsection{Motion and Clock Factors}
A motion factor is used to relate velocity to position. Specifically, the following constraint is introduced: When position and velocity are added to the estimated state, the difference in position between successive states is the integral of the average velocity between the states. The error function of the motion factor can be defined as

\begin{equation}
    \mathbf{e}_{\mathrm{M}}\left(\Delta \mathbf{x}_{i},\mathbf{v}_{i-1},\mathbf{v}_{i}\right)= \Delta \mathbf{x}_{i}-\Delta t\left(\frac{\mathbf{v}_{i-1}+\mathbf{v}_{i}}{2}\right) 
\end{equation}

Similarly, a clock factor establishes a relationship between the receiver clock and clock drift. Specifically, the following constraint is introduced: The average clock drift between states is equal to the difference in receiver clocks between states. The error function of the clock factor can be expressed as

\begin{equation}
    \mathbf{e}_{\mathrm{C}}\left(\Delta \mathbf{c}_{i},{d}_{i-1},{d}_{i}\right)=\Delta \mathbf{c}_{i}-\Delta t\left(\frac{{d}_{i-1}+{d}_{i}}{2}\right) 
\end{equation}

Notably, gtsam\_gnss provides a different type of clock factor that suppresses changes in the clock estimate. As the receiver clock changes continuously, unless a clock jump occurs, the clock change constraint can be applied even when clock drift is not included in the estimation state. This constraint ensures that the successive receiver clocks are equal and can be expressed as follows:

\begin{equation}
    \mathbf{e}_{\mathrm{C}}\left(\Delta \mathbf{c}_{i}\right)=\Delta \mathbf{c}_{i}
\end{equation}

These factors can help enhance the accuracy of position and clock estimation. In addition, gtsam\_gnss represents the position $\mathbf{x}$, velocity $\mathbf{v}$, clock $\mathbf{c}$, clock drift $d$, and ambiguity $\mathbf{B}$ using generic vector types provided in GTSAM. Therefore, the various generic factors in GTSAM can be applied directly to these estimated states.

\section{Application examples}
A key feature of gtsam\_gnss is that it provides not only different estimation states and factor combinations, but also analytical examples using these factors and actual GNSS observations. This section presents three use cases from the analysis examples in gtsam\_gnss, which demonstrate its value for GNSS analysis and processing.

\subsection{Example 1: Multipath Mitigation using Robust Error Model}
GNSS observations in urban environments are affected by ranging errors caused by multipath effects, which occur when GNSS signals are reflected or diffracted by obstacles such as buildings. In particular, NLOS multipath scenarios, where only reflected signals are received without any direct signals from satellites, represents a major source of error in GNSS positioning that is difficult to detect. Thus, NLOS multipath errors must be removed as outliers in GNSS positioning \cite{nlos_general,taro_multipath}.

\begin{figure}[t]
    \centerline{\includegraphics[width=88mm]{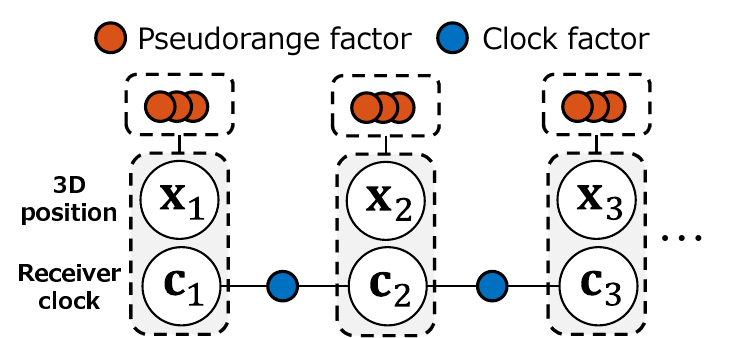}}
    \caption{Graph structure for Example 1. The GNSS pseudorange factor and clock factor, which suppress fluctuations in the receiver clock, are used to estimate the 3D position and receiver clock.}
    \label{fig2}
\end{figure}

In optimization, robust error models are typically used to reduce the effects of outliers \cite{slam_robust}. In this example, an M-estimator is applied to eliminate the multipath effects in GNSS pseudorange observations. Several robust error functions have been proposed for the M-estimator \cite{robustgnss}. In this work, the Huber function, which has been reported to be suitable for reducing GNSS multipath effects, is used \cite{HuberFGO,ion2021_taro}. Fig. 2 shows the graph structure for Example 1. This setup represents a simple single-point positioning model that uses GNSS pseudorange factors to estimate the 3D position and receiver clock. The clock factor is introduced as a constraint for clock estimation between states to limit the variation in the receiver clock. The optimization objective function is defined as follows:

\begin{IEEEeqnarray}{lCr}
    \widehat{\mathbf{X}}=\underset{\mathbf{X}}{\operatorname{argmin}} &\left( \sum_{i} \sum_{k}\left(\left\|\rho_{\mathrm{H}}\left(e_{\mathrm{P}}^k\left(\delta\mathbf{x}_i,\mathbf{c}_i\right)\right)\right\|_{\Omega_{\mathrm{P}}}^{2}\right) \right. \nonumber\\ &\left.+ \sum_{i} \left(\left\|\mathbf{e}_{\mathrm{C}}\left(\Delta \mathbf{c}_{i}\right)\right\|_{\Omega_{\mathrm{C}}}^{2}\right) \right)
 \end{IEEEeqnarray}

\noindent where $\rho_{\mathrm{H}}\left(\cdot\right)$ is the Huber function, which is a robust error model applied to the GNSS pseudorange factor. $\Omega_{\mathrm{P}}$ is an information matrix expressing the observation accuracy of the pseudorange, incorporating the satellite elevation angle–dependent pseudorange observation model \cite{matrtklib}. $\Omega_{\mathrm{C}}$ determines the clock continuity, set based on experimentally determined values. The objective function is optimized using the Levenberg--Marquardt optimizer in batch processing. 

The Precise Positioning Challenge (PPC) dataset \cite{ppc_dataset}, which contains GNSS data collected in a real-world environment, is used for the analysis. This dataset includes GNSS observations recorded in urban areas of Tokyo and Nagoya, Japan, using a vehicle-mounted GNSS receiver and antenna. Ground-truth data are acquired using a high-grade GNSS/INS integration system (Applanix POS LV). The GNSS data are acquired using Septentrio Mosaic-X5, with GPS, GLONASS, Galileo, BeiDou, and QZSS triple-frequency GNSS observations provided in the RINEX format. gtsam\_gnss incorporates a part of this PPC dataset.

The accuracies of position estimation with and without the M-estimator are compared. In the preprocessing stage, GNSS observations are rejected by thresholding the signal strength and satellite elevation angle to remove low-quality GNSS pseudoranges. Specifically, pseudorange observations with a received signal strength below 35 dB-Hz or a satellite elevation angle below 15° are rejected.

The position estimation results and 3D position estimation error for each method are shown in Fig. 3. In urban environments with dense buildings, large positioning errors occur owing to NLOS multipath errors, which cannot be completely eliminated by simple thresholding. However, when the M-estimator is used, these large positioning errors attributable to multipath effects are reduced. Fig. 4 shows the cumulative distribution of the 3D position error for each method. When the robust error model is used with the M-estimator, the positioning performance is considerably improved, even when using the same GNSS observations. This analysis demonstrates the effectiveness of using FGO and robust error models with GNSS observations for positioning in urban environments.

\begin{figure}[t]
    \centerline{\includegraphics[width=88mm]{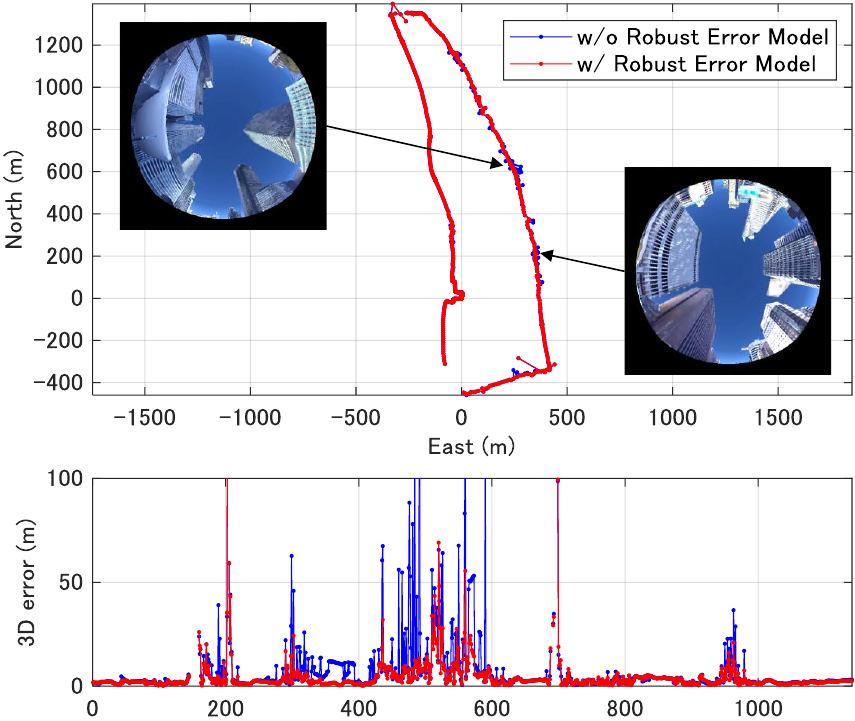}}
    \caption{Top: Position estimation results without (blue) and with (red) the robust error model. Bottom: 3D position estimation error for each method. Large position estimation errors occur in areas surrounded by buildings. However, using the robust error model helps enhance the position estimation accuracy.}
    \label{fig3}
\end{figure}

\begin{figure}[t]
    \centerline{\includegraphics[width=88mm]{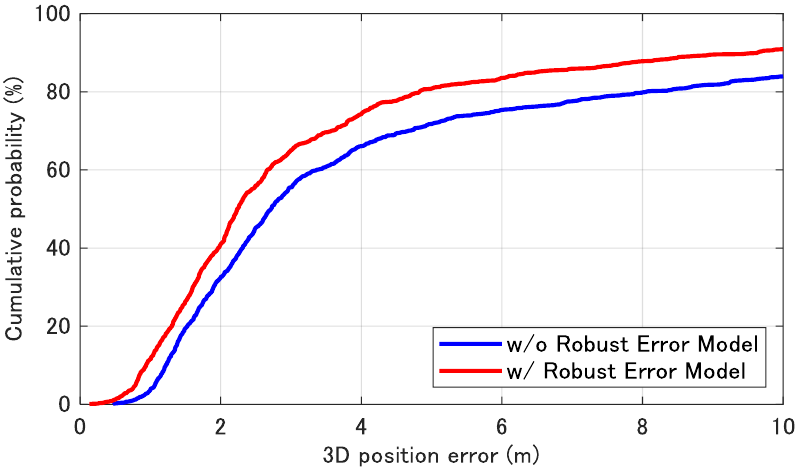}}
    \caption{Cumulative distribution of 3D position error. Using a robust error model (red) improves the position estimation accuracy.}
    \label{fig4}
\end{figure}

\subsection{Example 2: Carrier-Phase Ambiguity Resolution}
The success rate of integer ambiguity resolution of the GNSS carrier phase, which is essential for achieving centimeter-level positioning accuracy, depends on the accuracy and variance of the pre-estimated float ambiguities and 3D positions. In this example, the success rates of integer ambiguity resolution using the integer least-squares method with two different GNSS observation models within gtsam\_gnss are compared.

\begin{figure}[t]
    \centerline{\includegraphics[width=88mm]{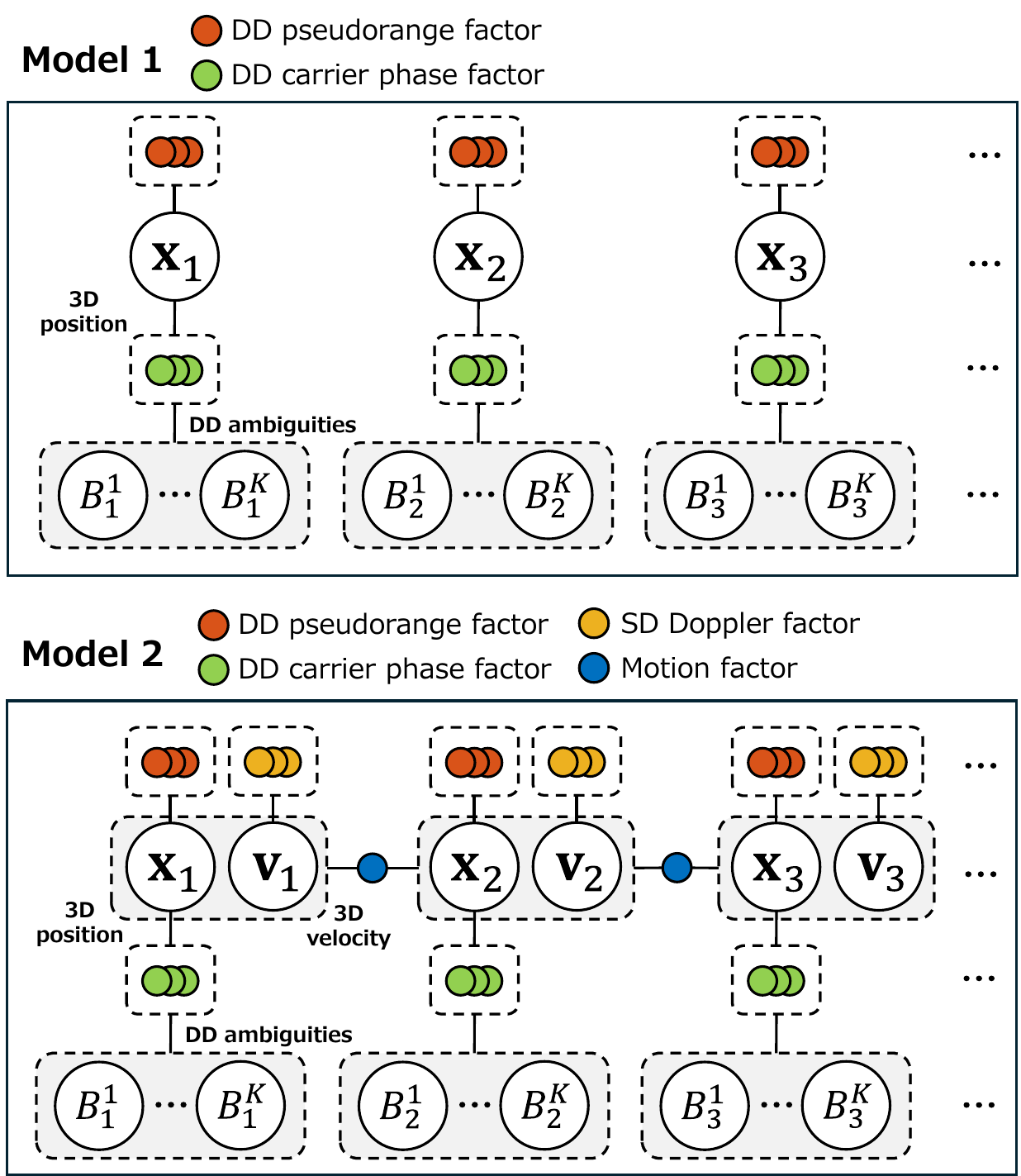}}
    \caption{Two graph structures for Example 2. Model 1 estimates the 3D position and DD ambiguity using DD pseudorange and DD carrier-phase factor. Model 2 extends Model 1 by adding velocity estimation using the Doppler factor.}
    \label{fig5}
\end{figure}

Fig. 5 shows the graph structure of the two models. Model 1 (top part of Fig. 5) estimates the 3D position and carrier-phase ambiguity from DD pseudorange and DD carrier-phase observations, representing a commonly used technique. The objective function of the FGO in Model 1 is expressed as follows:

\begin{IEEEeqnarray}{lCr}
    \widehat{\mathbf{X}}=\underset{\mathbf{X}}{\operatorname{argmin}} \left( \sum_{i} \sum_{k}\left\|e_{\mathrm{P}}^{kl}\left(\delta\mathbf{x}_i\right)\right\|_{\Omega_{\mathrm{P}}}^{2} \right. \nonumber\\ {} \left.+ \sum_{i} \sum_{k} \left\|e_{\mathrm{B}}^{kl}\left(\delta\mathbf{x}_{i}, \nabla \Delta B_{i}^{kl}\right)\right\|_{\Omega_{\mathrm{B}}}^{2} \right)
 \end{IEEEeqnarray}

The graph structure of Model 2 is shown in the bottom part of Fig. 5. Model 2 extends Model 1 by incorporating Doppler observations to estimate velocity and using velocity constraints between the 3D position states. Doppler observations are known to be more robust to multipath effects than pseudorange observations, and by using a motion factor to relate the position and velocity between states, the positioning and ambiguity estimation accuracies can be improved. The objective function of FGO in Model 2 can be expressed as

\begin{IEEEeqnarray}{lCr}
    \widehat{\mathbf{X}}=&\underset{\mathbf{X}}{\operatorname{argmin}} \left( \sum_{i} \sum_{k}\left\|e_{\mathrm{P}}^{kl}\left(\delta\mathbf{x}_i\right)\right\|_{\Omega_{\mathrm{P}}}^{2} + \sum_{i} \sum_{k}\left\|e_{\mathrm{D}}^{kl}\left(\delta\mathbf{v}_i\right)\right\|_{\Omega_{\mathrm{D}}}^{2} \right. \nonumber\\  & \left. + \sum_{i} \sum_{k} \left\|e_{\mathrm{B}}^{kl}\left(\delta\mathbf{x}_{i}, \nabla \Delta B_{i}^{kl}\right)\right\|_{\Omega_{\mathrm{B}}}^{2} \right. \nonumber\\ & \left. + \sum_{i} \left\|\mathbf{e}_{\mathrm{M}}\left(\Delta \mathbf{x}_{i},\mathbf{v}_{i-1},\mathbf{v}_{i}\right)\right\|_{\Omega_{\mathrm{M}}}^{2}\right)
 \end{IEEEeqnarray}

\noindent where the inter-satellite difference Doppler observations are used to eliminate the receiver clock drift present in the Doppler observations.

These two models are compared to assess the contribution of Doppler observations to ambiguity resolution. The same GNSS dataset as that used in Example 1 is used for the analysis. The LAMBDA method \cite{lambda} is used to estimate the integer ambiguity. The float solution, along with the float ambiguity estimated by FGO and corresponding variance, is input into the LAMBDA method. The integer ambiguity is determined through a search process and validated using the ratio test \cite{ratiotest}. The threshold for the ratio test is set as 2.0.

Fig. 6 shows the ratio of integer ambiguities output by the LAMBDA method for each method. When the Doppler factor is included, the accuracy of the float solution improves, and the ratio of the LAMBDA method is enhanced. Consequently, the success rate of integer ambiguity resolution (fixed rate) increases. The fixed rates for Models 1 and 2 are 40.8\%, and 51.4\%. These results confirm that the Doppler factor positively contributes to integer ambiguity resolution.

Fig. 7 shows the cumulative distribution of the 3D position error for each model. The use of the Doppler factor improves the centimeter-level position estimation accuracy. Moreover, the overall estimation accuracy is enhanced owing to the improved estimation accuracy of the float solution. These results show that the combination of factors used considerably affects the success rate of ambiguity resolution of the carrier phase.

\begin{figure}[t]
    \centerline{\includegraphics[width=88mm]{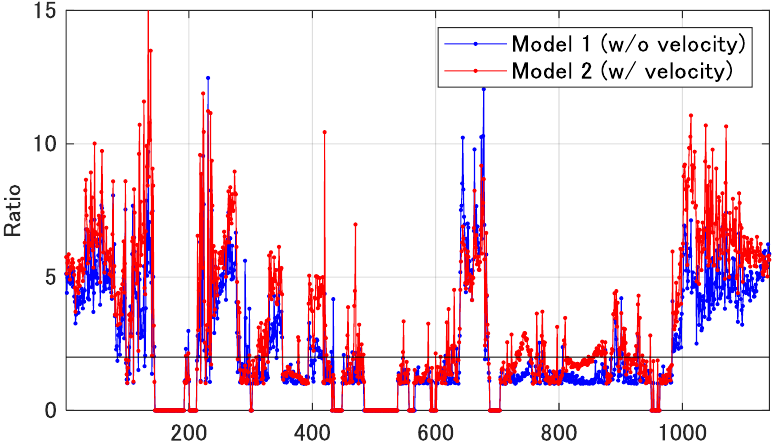}}
    \caption{Ratio test for integer solutions of the LAMBDA method, computed using two types of factor graph models. The black line indicates the threshold. Compared with Model 1 (blue), Model 2 (red), which incorporates the velocity factor, exhibits an enhanced ratio).
    }
    \label{fig6}
\end{figure}

\begin{figure}[t]
    \centerline{\includegraphics[width=88mm]{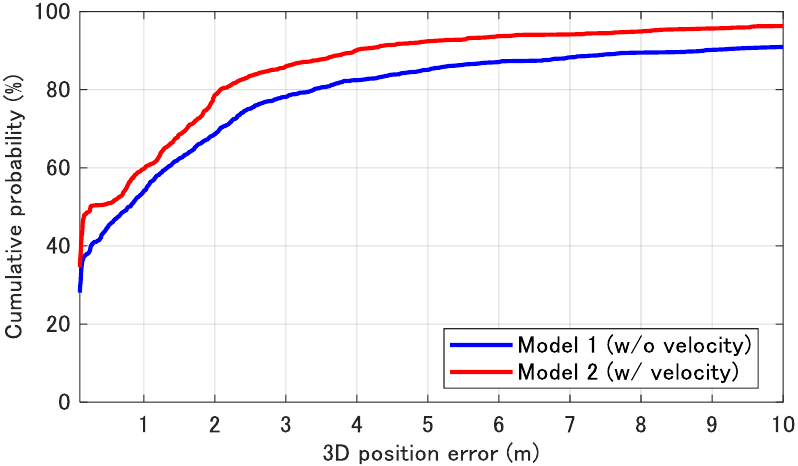}}
    \caption{Cumulative distribution of 3D position estimation errors computed using two types of factor graph models. Model 2 (red), which uses velocity factors, exhibits improved position estimation accuracy compared with Model 1 (blue).}
    \label{fig7}
\end{figure}

\subsection{Example 3: Smartphone Location Estimation in the Smartphone Decimeter Challenge (SDC)}
An API for accessing raw GNSS observations from Android smartphones was released in 2016 \cite{api}, paving the way for high-precision positioning using smartphones. To further support the development of high-precision positioning technologies for smartphones, Google hosted the SDC in 2021 \cite{gsdc}. The SDC provided raw GNSS and IMU data collected from in-vehicle smartphones \cite{gsd} and tasked participants to accurately estimate the smartphone positions along predefined routes. This challenge was held annually from 2021 to 2023, during which we developed a highly accurate positioning method for smartphones using FGO that achieved first place in 2021 \cite{gsdc2021_taro}, first place in 2022 \cite{gsdc2022_taro,gsdc2022_sensors}, and second place \cite{gsdc2023_taro} in 2023. In addition, decimeter-level positioning accuracy was achieved in SDC2023, and the corresponding technique was released as an open-source code \cite{gsdc2023_oss}.

\begin{figure}[t]
    \centerline{\includegraphics[width=88mm]{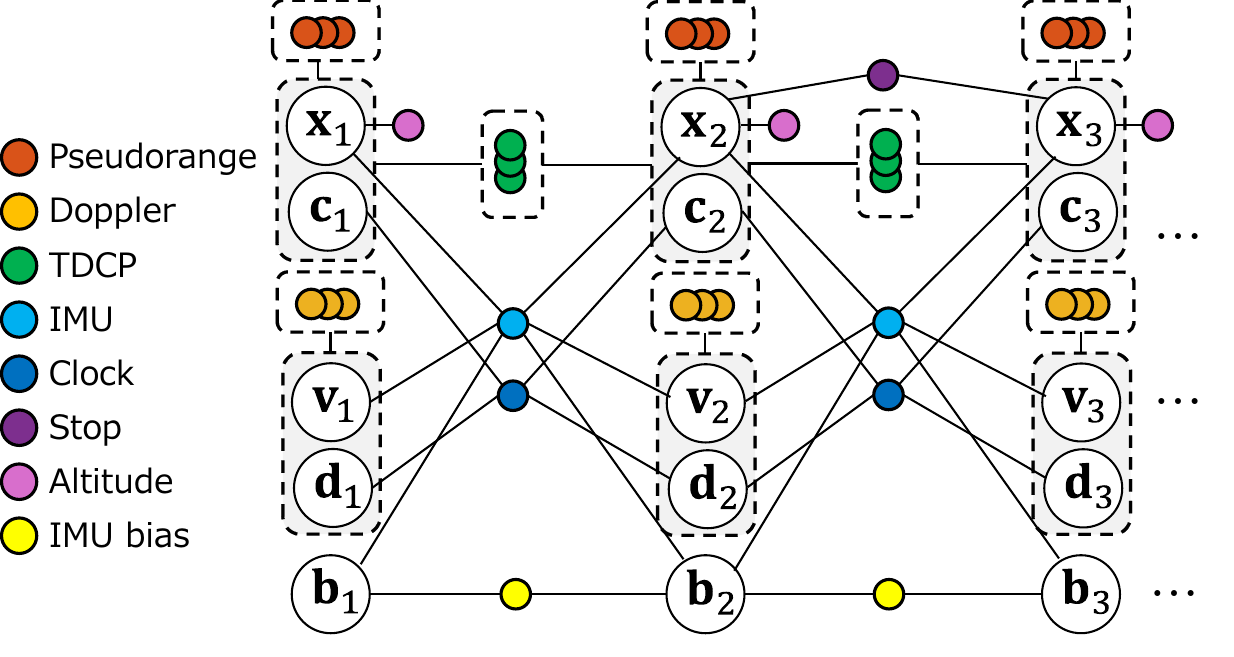}}
    \caption{Factor graph model for estimating smartphone location, used in SDC2023. The approach uses eight factors to estimate the location from GNSS and IMU observations.}
    \label{fig8}
\end{figure}

This analysis demonstrates the achievement of decimeter-level accuracy using gtsam\_gnss on a smartphone. The main challenge associated with GNSS positioning on a smartphone is the low quality of GNSS observations owing to the limitations of the smartphone antenna. Compared with commercial GNSS receivers, smartphone antennas are more susceptible to multipath effects, and the signal reception level is lower, resulting in increased observation noise \cite{phone3}. The proposed method improves the position estimation accuracy by integrating the IMU observations of the smartphone into a factor graph, along with GNSS observation factors. Moreover, by adding various factors, such as elevation constraints and vehicle-stop constraints, the position estimation performance is successfully improved.

Fig. 8 shows the factor graph used for high-precision positioning of smartphones used in SDC2023. The framework incorporates eight factors: pseudorange, Doppler, TDCP, IMU, clock, stop, altitude, and IMU bias factors. Using this graph structure, the position, attitude, velocity, clock and clock drift, and IMU bias of the smartphone are determined through optimization. Details of the processing method can be found in the SDC2023 solution paper \cite{gsdc2023_taro}.

Fig. 9 presents the horizontal position estimation errors for smartphones on highways and vehicles on city streets, representative of the driving environments considered in the SDC. The results are compared with the position estimates based on the least-squares method applied to pseudorange computations within the smartphone. The accuracy of position estimates based on the smartphone is approximately 3.2 m in the highway environment, where signals are unobstructed. In contrast, the proposed method achieves a considerably higher accuracy of 0.39 m. In the city environment, the positioning accuracy of the conventional method reduces to approximately 4.7 m owing to the deterioration of the GNSS signal quality. However, the proposed method using FGO still estimates positions with a high accuracy of 0.61 m.

\begin{figure}[t]
    \centerline{\includegraphics[width=88mm]{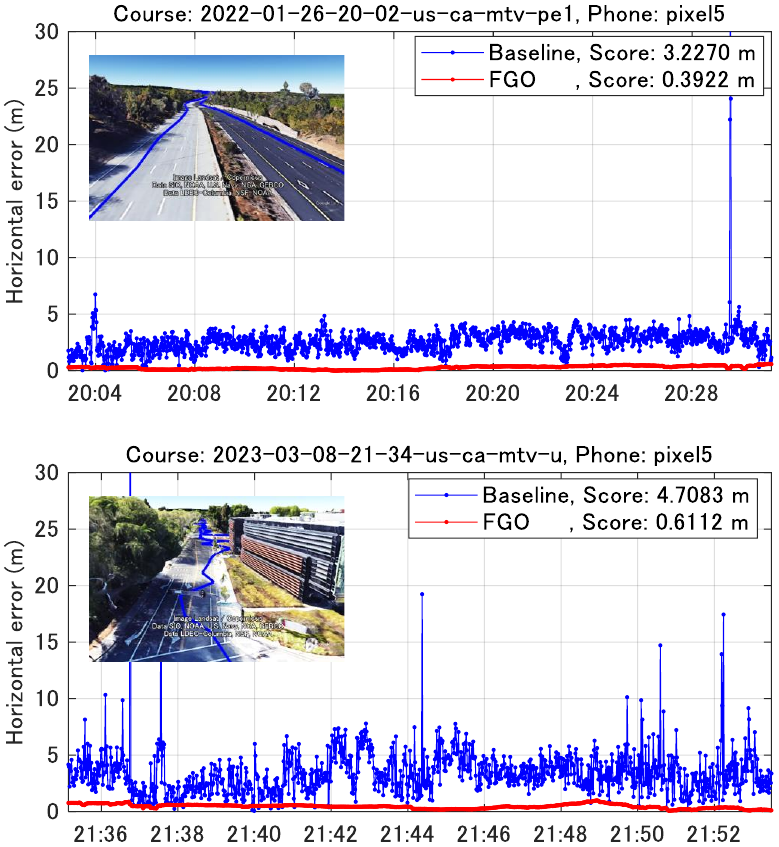}}
    \caption{Example 3: SDC2023 data analysis. Top: Highway driving data. Bottom: City driving data. Compared with the position calculated within the receiver (blue), the FGO (red) estimates positions with high accuracy and reduced errors.}
    \label{fig9}
\end{figure}

Table 1 lists the root mean square error, 50th percentile error, and 95th percentile error, representing statistical values of the horizontal error associated with the proposed position-estimation method for 40 drives included in the SDC2023 training dataset. The table also lists the average values of the 50th and 95th percentile errors, which are the evaluation metrics of the SDC. The proposed method using FGO achieves a higher positioning accuracy compared with the conventional receiver-based method. Notably, the proposed method achieves decimeter-level positioning accuracy using GNSS and IMU observations from a smartphone.

\begin{table}[]
    \centering
    \caption{Positioning error statistics for 40 runs in the SDC2023 training dataset.}
    \begin{tabular}{lcccc}
    \hline
             & \begin{tabular}[c]{@{}c@{}}RMS \\ error m\end{tabular} & \begin{tabular}[c]{@{}c@{}}50th percentile\\ error m\end{tabular} & \begin{tabular}[c]{@{}c@{}}95th percentile\\ error m\end{tabular} & \begin{tabular}[c]{@{}c@{}}SDC score\\ m\end{tabular} \\ \hline
    Baseline & 3.890                                                  & 2.644                                                             & 6.867                                                             & 4.756                                                 \\
    FGO      & 0.587                                                  & 0.495                                                             & 0.967                                                             & 0.731                                                 \\ \hline
    \end{tabular}
\end{table}
\section{Conclusion}
This paper introduces a simple GNSS factor graph optimization package that can be easily applied to GNSS research and development. This package, which has been released as an open-source project, describes the factors of GNSS observation by separating the preprocessing and optimization processes of GNSS observations. Consequently, the provided factors have general-purpose inputs, and different combinations of factors can be selected for specific FGO problems.

A key advantage of the gtsam\_gnss package is that it provides concrete examples of graph optimization using different types of GNSS observations. This paper highlights three application examples: robust error modeling, estimation of integer ambiguity of the carrier phase, and integration of GNSS and IMU observations from smartphones. Excellent state estimation performance is achieved in all cases, demonstrating the effectiveness of the proposed GNSS FGO package.

The gtsam\_gnss remains under active development, with future work aimed at introducing additional factors related to GNSS observations. Future updates will also include examples demonstrating how to use these factors, and source code will be provided to support further learning and development in the GNSS and FGO domains.

\bibliographystyle{IEEEtran}
\balance
\bibliography{IEEEabrv,IONPLANS2025}

\begin{thebibliography}{10}
\providecommand{\url}[1]{#1}
\csname url@samestyle\endcsname
\providecommand{\newblock}{\relax}
\providecommand{\bibinfo}[2]{#2}
\providecommand{\BIBentrySTDinterwordspacing}{\spaceskip=0pt\relax}
\providecommand{\BIBentryALTinterwordstretchfactor}{4}
\providecommand{\BIBentryALTinterwordspacing}{\spaceskip=\fontdimen2\font plus
\BIBentryALTinterwordstretchfactor\fontdimen3\font minus \fontdimen4\font\relax}
\providecommand{\BIBforeignlanguage}[2]{{%
\expandafter\ifx\csname l@#1\endcsname\relax
\typeout{** WARNING: IEEEtran.bst: No hyphenation pattern has been}%
\typeout{** loaded for the language `#1'. Using the pattern for}%
\typeout{** the default language instead.}%
\else
\language=\csname l@#1\endcsname
\fi
#2}}
\providecommand{\BIBdecl}{\relax}
\BIBdecl

\bibitem{urbannav}
L.-T. Hsu, N.~Kubo, W.~Chen, Z.~Liu, T.~Suzuki, and J.~Meguro, ``{UrbanNav:An Open-sourced Multisensory Dataset for Benchmarking Positioning Algorithms Designed for Urban Areas},'' in \emph{34th International Technical Meeting of the Satellite Division of the Institute of Navigation, ION GNSS+ 2021}.\hskip 1em plus 0.5em minus 0.4em\relax Institute of Navigation, 2021.

\bibitem{nlos_general}
P.~D. Groves, Z.~Jiang, M.~Rudi, and P.~Strode, ``{A Portfolio Approach to NLOS and Multipath Mitigation in Dense Urban Areas},'' in \emph{26th International Technical Meeting of the Satellite Division of the Institute of Navigation, ION GNSS 2013}, vol.~4, 2013, pp. 3231--3247.

\bibitem{gpsins}
P.~D. Groves, \emph{{Principles of GNSS, Inertial, and Multisensor Integrated Navigation Systems}}.\hskip 1em plus 0.5em minus 0.4em\relax Norwood, MA, USA: Artech House, 2013.

\bibitem{gnss_handbook}
P.~J. Teunissen and O.~Montenbruck, \emph{{Springer Handbook of Global Navigation Satellite Systems}}.\hskip 1em plus 0.5em minus 0.4em\relax Springer, 2017, vol.~10.

\bibitem{gnssfgo_2012}
N.~S{\"{u}}nderhauf, M.~Obst, G.~Wanielik, and P.~Protzel, ``{Multipath mitigation in GNSS-based localization using robust optimization},'' in \emph{IEEE Intelligent Vehicles Symposium}, 2012, pp. 784--789.

\bibitem{gnssfgo_survey}
\BIBentryALTinterwordspacing
C.~Taylor and J.~Gross, ``Factor graphs for navigation applications: A tutorial,'' \emph{NAVIGATION: Journal of the Institute of Navigation}, vol.~71, no.~3, 2024. [Online]. Available: \url{https://navi.ion.org/content/71/3/navi.653}
\BIBentrySTDinterwordspacing

\bibitem{gnssfgo_kf}
\BIBentryALTinterwordspacing
W.~Wen, T.~Pfeifer, X.~Bai, and L.-T. Hsu, ``Factor graph optimization for gnss/ins integration: A comparison with the extended kalman filter,'' \emph{NAVIGATION}, vol.~68, no.~2, pp. 315--331, 2021. [Online]. Available: \url{https://onlinelibrary.wiley.com/doi/abs/10.1002/navi.421}
\BIBentrySTDinterwordspacing

\bibitem{fgo_general1}
F.~R. Kschischang, B.~J. Frey, and H.~A. Loeliger, ``{Factor graphs and the sum-product algorithm},'' \emph{IEEE Transactions on Information Theory}, vol.~47, no.~2, pp. 498--519, 2001.

\bibitem{fgo_book}
F.~Dellaert, M.~Kaess \emph{et~al.}, ``Factor graphs for robot perception,'' \emph{Foundations and Trends{\textregistered} in Robotics}, vol.~6, no. 1-2, pp. 1--139, 2017.

\bibitem{orbslam}
R.~Mur-Artal, J.~M.~M. Montiel, and J.~D. Tardos, ``Orb-slam: A versatile and accurate monocular slam system,'' \emph{IEEE transactions on robotics}, vol.~31, no.~5, pp. 1147--1163, 2015.

\bibitem{cartographer}
W.~Hess, D.~Kohler, H.~Rapp, and D.~Andor, ``Real-time loop closure in 2d lidar slam,'' in \emph{2016 IEEE international conference on robotics and automation (ICRA)}.\hskip 1em plus 0.5em minus 0.4em\relax IEEE, 2016, pp. 1271--1278.

\bibitem{robustgnss}
R.~M. Watson and J.~N. Gross, ``{Robust Navigation in GNSS Degraded Environment Using Graph Optimization},'' in \emph{30th International Technical Meeting of the Satellite Division of the Institute of Navigation, ION GNSS 2017}, vol.~5.\hskip 1em plus 0.5em minus 0.4em\relax Institute of Navigation, 2017, pp. 2906--2918.

\bibitem{graphgnsslib}
W.~Wen and L.-T. Hsu, ``Towards robust gnss positioning and real-time kinematic using factor graph optimization,'' in \emph{2021 IEEE International Conference on Robotics and Automation (ICRA)}, 2021, pp. 5884--5890.

\bibitem{g2o}
R.~K{\"u}mmerle, G.~Grisetti, H.~Strasdat, K.~Konolige, and W.~Burgard, ``g2o: A general framework for graph optimization,'' in \emph{2011 IEEE international conference on robotics and automation}.\hskip 1em plus 0.5em minus 0.4em\relax IEEE, 2011, pp. 3607--3613.

\bibitem{ceres}
S.~Agarwal, K.~Mierle \emph{et~al.}, ``Ceres solver: Tutorial \& reference,'' \emph{Google Inc}, vol.~2, no.~72, p.~8, 2012.

\bibitem{gtsam}
F.~Dellaert, ``Factor graphs and gtsam: A hands-on introduction,'' \emph{Georgia Institute of Technology, Tech. Rep}, vol.~2, p.~4, 2012.

\bibitem{hdlslam}
K.~Koide, J.~Miura, and E.~Menegatti, ``A portable three-dimensional lidar-based system for long-term and wide-area people behavior measurement,'' \emph{International Journal of Advanced Robotic Systems}, vol.~16, no.~2, p. 1729881419841532, 2019.

\bibitem{liosam}
T.~Shan, B.~Englot, D.~Meyers, W.~Wang, C.~Ratti, and R.~Daniela, ``Lio-sam: Tightly-coupled lidar inertial odometry via smoothing and mapping,'' in \emph{IEEE/RSJ International Conference on Intelligent Robots and Systems (IROS)}.\hskip 1em plus 0.5em minus 0.4em\relax IEEE, 2020, pp. 5135--5142.

\bibitem{gnssfgo}
H.~Zhang, C.-C. Chen, H.~Vallery, and T.~D. Barfoot, ``Gnss/multi-sensor fusion using continuous-time factor graph optimization for robust localization,'' \emph{IEEE Transactions on Robotics}, 2024.

\bibitem{posgo}
Z.~Li, J.~Guo, and Q.~Zhao, ``Posgo: an open-source software for gnss pseudorange positioning based on graph optimization,'' \emph{GPS Solutions}, vol.~27, no.~4, p. 187, 2023.

\bibitem{gicilib}
C.~Chi, X.~Zhang, J.~Liu, Y.~Sun, Z.~Zhang, and X.~Zhan, ``Gici-lib: A gnss/ins/camera integrated navigation library,'' \emph{IEEE Robotics and Automation Letters}, 2023.

\bibitem{gvins}
S.~Cao, X.~Lu, and S.~Shen, ``Gvins: Tightly coupled gnss--visual--inertial fusion for smooth and consistent state estimation,'' \emph{IEEE Transactions on Robotics}, vol.~38, no.~4, pp. 2004--2021, 2022.

\bibitem{matrtklib}
\BIBentryALTinterwordspacing
T.~Suzuki, ``Matrtklib: a matlab‑based open-source software for gnss data processing.'' [Online]. Available: \url{https://github.com/taroz/MatRTKLIB}
\BIBentrySTDinterwordspacing

\bibitem{vel1}
L.~Serrano, D.~Kim, R.~B. Langley, K.~Itani, and M.~Ueno, ``{A GPS velocity sensor: How accurate can it be? - A first look},'' in \emph{Proceedings of the National Technical Meeting, Institute of Navigation}, vol. 2004, 2004, pp. 875--885.

\bibitem{td1}
P.~Freda, A.~Angrisano, S.~Gaglione, and S.~Troisi, ``{Time-differenced carrier phases technique for precise GNSS velocity estimation},'' \emph{GPS Solutions}, vol.~19, no.~2, pp. 335--341, 2015.

\bibitem{ral2022_taro}
T.~Suzuki, ``Gnss odometry: Precise trajectory estimation based on carrier phase cycle slip estimation,'' \emph{IEEE Robotics and Automation Letters}, vol.~7, no.~3, pp. 7319--7326, 2022.

\bibitem{fgo_tdcp}
\BIBentryALTinterwordspacing
Y.~Jiang, Y.~Gao, W.~Ding, and Y.~Gao, ``Gnss precise positioning for smartphones based on the integration of factor graph optimization and solution separation,'' \emph{Measurement}, vol. 203, p. 111924, 2022. [Online]. Available: \url{https://www.sciencedirect.com/science/article/pii/S0263224122011204}
\BIBentrySTDinterwordspacing

\bibitem{ar_review}
D.~Kim and R.~B. Langley, ``Gps ambiguity resolution and validation: methodologies, trends and issues,'' in \emph{Proceedings of the 7th GNSS Workshop--International Symposium on GPS/GNSS, Seoul, Korea}, vol.~30, no. 2.12, 2000.

\bibitem{taro_multipath}
T.~Suzuki, ``Mobile robot localization with gnss multipath detection using pseudorange residuals,'' \emph{Advanced Robotics}, vol.~33, no.~12, pp. 602--613, 2019.

\bibitem{slam_robust}
N.~S{\"{u}}nderhauf and P.~Protzel, ``{Towards a robust back-end for pose graph SLAM},'' in \emph{IEEE International Conference on Robotics and Automation}, 2012, pp. 1254--1261.

\bibitem{HuberFGO}
X.~Bai, W.~Wen, and L.-T. Hsu, ``Time-correlated window-carrier-phase-aided gnss positioning using factor graph optimization for urban positioning,'' \emph{IEEE Transactions on Aerospace and Electronic Systems}, vol.~58, no.~4, pp. 3370--3384, 2022.

\bibitem{ion2021_taro}
T.~Suzuki, ``Robust vehicle positioning in multipath environments based on graph optimization,'' in \emph{Proceedings of the 34th International Technical Meeting of the Satellite Division of The Institute of Navigation (ION GNSS+ 2021)}, 2021, pp. 4223--4233.

\bibitem{ppc_dataset}
\BIBentryALTinterwordspacing
T.~Suzuki, ``Precise positioning challenge dataset.'' [Online]. Available: \url{https://github.com/taroz/PPC-Dataset}
\BIBentrySTDinterwordspacing

\bibitem{lambda}
P.~J. Teunissen, ``Least-squares estimation of the integer gps ambiguities,'' in \emph{Invited lecture, section IV theory and methodology, IAG general meeting, Beijing, China}, 1993, pp. 1--16.

\bibitem{ratiotest}
S.~Verhagen and P.~J. Teunissen, ``The ratio test for future gnss ambiguity resolution,'' \emph{GPS solutions}, vol.~17, pp. 535--548, 2013.

\bibitem{api}
T.~E. Humphreys, M.~Murrian, F.~Van~Diggelen, S.~Podshivalov, and K.~M. Pesyna, ``On the feasibility of cm-accurate positioning via a smartphone's antenna and gnss chip,'' in \emph{2016 IEEE/ION position, location and navigation symposium (PLANS)}.\hskip 1em plus 0.5em minus 0.4em\relax IEEE, 2016, pp. 232--242.

\bibitem{gsdc}
\BIBentryALTinterwordspacing
G.~M. Fu, M.~Khider, and F.~van Diggelen. (2021) Google smartphone decimeter challenge. [Online]. Available: \url{https://www.kaggle.com/c/google-smartphone-decimeter-challenge/}
\BIBentrySTDinterwordspacing

\bibitem{gsd}
G.~M. Fu, M.~Khider, and F.~van Diggelen, ``Android raw gnss measurement datasets for precise positioning,'' in \emph{Proceedings of the 33rd International Technical Meeting of the Satellite Division of The Institute of Navigation (ION GNSS+ 2020)}, 2020, pp. 1925--1937.

\bibitem{gsdc2021_taro}
T.~Suzuki, ``First place award winner of the smartphone decimeter challenge: global optimization of position and velocity by factor graph optimization,'' in \emph{Proceedings of the 34th International Technical Meeting of the Satellite Division of The Institute of Navigation (ION GNSS+ 2021)}, 2021, pp. 2974--2985.

\bibitem{gsdc2022_taro}
T.~Suzuki, ``1st place winner of the smartphone decimeter challenge: two-step optimization of velocity and position using smartphone’s carrier phase observations,'' in \emph{Proceedings of the 35th International Technical Meeting of the Satellite Division of The Institute of Navigation (ION GNSS+ 2022)}, 2022, pp. 2276--2286.

\bibitem{gsdc2022_sensors}
T.~Suzuki, ``Precise position estimation using smartphone raw gnss data based on two-step optimization,'' \emph{Sensors}, vol.~23, no.~3, p. 1205, 2023.

\bibitem{gsdc2023_taro}
T.~Suzuki, ``Second place winner of the smartphone decimeter challenge: An open-source factor graph optimization package for gnss and imu integration in smartphones,'' in \emph{Proceedings of the 37th International Technical Meeting of the Satellite Division of The Institute of Navigation (ION GNSS+ 2024)}, 2024, pp. 2703--2713.

\bibitem{gsdc2023_oss}
\BIBentryALTinterwordspacing
T.~Suzuki, ``gsdc2023.'' [Online]. Available: \url{https://github.com/taroz/gsdc2023}
\BIBentrySTDinterwordspacing

\bibitem{phone3}
G.~Li and J.~Geng, ``Characteristics of raw multi-gnss measurement error from google android smart devices,'' \emph{GPS Solutions}, vol.~23, no.~3, pp. 1--16, 2019.

\end{thebibliography}
\end{document}